\documentclass[12pt]{article}
\usepackage[a4paper, total={6in, 8in}]{geometry}

\usepackage{setspace,graphicx,amssymb,amsmath,latexsym,amsfonts,amscd,amsthm,multirow,ctable,mathdots,caption,array}
\usepackage{fancyhdr,tabularx,cite,mathrsfs}
\usepackage{authblk}
\usepackage{hyperref}

\usepackage{times}

\newcommand{\CC}{\mathbb{C}} 

\DeclareMathOperator*{\argmax}{arg\,max}
\DeclareMathOperator{\Tr}{Tr}

\begin{document}

\title{Entangled q-Convolutional Neural Nets}


\author{Vassilis~Anagiannis\thanks{vassilis\textunderscore 1991@yahoo.gr}~\thanks{Institute of Physics, University of Amsterdam, Amsterdam, the Netherlands} ~~ Miranda C. N.  Cheng\thanks{mcheng@uva.nl}~\protect\footnotemark[2]}
\date{}


\maketitle

\vspace{-5pt}
\begin{abstract}
We introduce a machine learning model, the {\em q-CNN model}, sharing key features with convolutional neural networks and admitting a tensor network description. As examples, we apply q-CNN to the MNIST and Fashion MNIST classification tasks. We explain how the network associates a quantum state to each classification label, and study the entanglement structure of these network states. 
In both our experiments on the MNIST and Fashion-MNIST datasets, we observe a distinct increase in both the left/right as well as the up/down bipartition entanglement entropy during training as the network learns the fine features of the data. More generally, we observe a universal negative correlation between the value of the entanglement entropy and the value of the cost function, suggesting that the network needs to learn  the entanglement structure in order the perform the task accurately. This supports the possibility of exploiting the entanglement structure as a guide to design the machine learning algorithm suitable for given tasks.
\end{abstract}


\section{Introduction}

Convolutional neural networks (CNNs)  have seen remarkable successes in various applications. At the same time there are tasks with similar descriptions that can nevertheless not be solved with a CNN architecture\footnote{See, for instance, \cite{vortices} for an example in physics.}. Moreover, it is not always transparent what choices of hyperparameters work the best, and why. More generally, we do not always have a precise explanation of why certain choices of 
machine learning architectures and hyperparameters work and do not work for a given task. 
This lack of a precise understanding is related to the curse of dimensionality which prevents an explicit analysis. That said, the data of a given problem typically lie in a high co-dimensional subspace. For instance, a typical point in the configuration space of all possible $N$-pixel grayscale pictures resembles a ``white noise'' image, and looks nothing like a picture encountered in the relevant data set.

This is very reminiscent of the situation in quantum many-body systems. The high dimensionality of the Hilbert space of quantum states makes it hard to find the desired state (e.g. ground state of the given Hamiltonian) explicitly. Tensor network \cite{bridgeman2017hand, biamonte2017tensor} is one of the most popular tools utilised in many-body quantum physics to overcome this problem. 
Abstractly speaking, they provide a way to approximate high-order tensors in terms of lower-order tensors, and by doing so  greatly reduce the parameters needed to describe the relevant quantum states, circumventing the curse of dimensionality.
This is possible because the physically relevant states lie in a tiny ``corner of the Hilbert space". One can quantify this using the {\em entanglement entropy} (EE) of a quantum state with respect to a bipartition of the system, which measures the degree to which the quantum state is entangled between two subsystems. While a typical element of the Hilbert space has an entanglement entropy which scales like the volume of the sub-region, the physically relevant states tend to have entanglement entropies that scale like the boundary area (possibly with logarithmic corrections) of the sub-region.
At the same time, the entanglement structure of a quantum state is precisely what constrains how effectively it could be approximated by a given tensor network architecture. See \cite{eisert2013entanglement,eisert2010colloquium} for a review.

The analogy between quantum many-body systems and machine learning prompts the following  questions. 
Could we have a similar theoretical understanding in the context of machine learning architectures on how effective they are for a given task? Could we also understand the subspace of relevant data with similar tools as those used in the study of quantum many-body systems? 
Moreover, one might also hope that the analogy between quantum many-body systems and machine learning architectures can help the developments of natural and effective quantum machine learning architectures \cite{cong2019quantum, bondesan2020quantum, huggins2019towards}.

Inspired by these questions, there have been increasing efforts to build a bridge between the two fields of machine learning and quantum many-body systems, and in particular tensor networks \cite{carrasquilla2017machine, amin2018quantum,novikov2016exponential, cohen2016inductive,cohen2016convolutional, liu2018machine,li2015experimental, cichocki2016tensor,cichocki2017tensor, han2018unsupervised,carleo2017solving, chen2018equivalence,huang2017neural, glasser2018neural, cohen2016expressive, levine2017deep, pestun2017tensor}.
In this work, we continue to strengthen this bridge, focusing on convolutional neural networks (CNNs).
Specifically, we build a CNN-like architecture, which we call q-CNN, which admits a description as a tensor network. In particular, our architecture has the same weight sharing property as the usual CNN, and as a result the number of parameters grows only logarithmically with the system size. 
Subsequently, we apply q-CNN to the  classification tasks on the MNIST and Fashion-MNIST datasets, obtaining maximum test accuracy $97\%$ and $89\%$ respectively, comparable to \cite{efthymiou2019tensornetwork}.
With the confidence that our architecture has satisfactory performance, we then go ahead and explore its quantum mechanical properties. 
As mentioned before, a crucial probe of the qualitative features of a quantum states is its entanglement entropies. 
We compute the entanglement entropies of the network quantum states, with respect to bipartitions of the configuration space corresponding to the left/right and up/down partitions of an image.
In both our experiments on the MNIST and Fashion-MNIST datasets, we initialize the network in a random way which renders a quantum state with no particular entanglement structure. Subsequently, we observe a distinct increase in entanglement entropy during training as the network learns the fine features of the data.  More generally, we observe a  negative correlation between the value of the entanglement entropy and the value of the cost function, across different initialisations and choices of hyperparameters of the network. This constitutes convincing evidence that one of the structures of the data that the network needs to ``learn'' is the entanglement structure. It can also be read out that the entanglement needed for the (Fashion-) MNIST classification tasks is low, which could be viewed as a  ``justification'' why a simple CNN is capable of performing these tasks.

\subsection{Related Work}
The q-CNN architecture discussed in this work is based on the theoretical architecture, named deep convolutional arithmetic circuit,  introduced in \cite{cohen2016expressive, levine2017deep}. 
The product pooling proposed in \cite{cohen2016expressive, levine2017deep} presents a challenge in training the network described above, since it can easily lead to numerical instabilities such as underflow or overflow. We would like to  render the network trainable in practice, and do so without spoiling the analogy to quantum many-body systems. In \cite{sharir2016tensorial} this architecture was trained using simnets \cite{cohen2016deep},  circumventing the numerical instability of product pooling by performing the calculations in logarithmic space. In q-CNN, we instead introduce additional batch normalisation layers and incorporate them into the tensor description of the network. 

As opposed to other works aiming to study and/or practice machine learning in ways inspired by  quantum many-body physics  
\cite{stoudenmire2016supervised,stoudenmire2018learning,efthymiou2019tensornetwork}, we train the network as a usual neural network with Pytorch instead of 
optimization schemes commonly used for tensor networks, such as DMRG.
Also note that the number of parameters of our network grows merely logarithmically with the size of the image, as opposed to linearly as in the aforementioned approaches, since we retain the weight sharing feature of the usual convolutional neural network in our architecture. 
In \cite{liu2019machine}, the entanglement entropy of the final trained network was computed for a very different architecture directly related to tensor networks, but not its evolution during training and the correlation between entanglement and accuracy. 
In \cite{liu2018entanglement} and \cite{levine2017deep}, the possibility was  mentioned that the requirement of being capable to accommodate the entanglement could be used to guide the design of the network. 
We note that the values of the entanglement entropy in our experiments on the MNIST and F-MNIST datasets is of the order of $\log2$. As a result, the entanglement is unlikely to be a bottleneck of the network performance for such tasks.

\section{The q-CNN Architecture}
\label{sec:architecture} 

In this section we describe the architecture of q-CNN. 

Consider a grayscale image with $N$ pixels, corresponding to a point in the configuration space 
${\bf x} = (x_1, \dots, x_N) \in [0,1]^{\otimes N}$. For simplicity we have flattened the 2d image into a chain. 
We will take $N=2^{L}$ with integral $L$, using padding if necessary.  
The first layer of the q-CNN is a (non-linear)  representation layer: 
\begin{gather}\begin{split}
{\rm Rep}: ~~ &  [0,1]^{\otimes N} \to ({\mathbb R}^{d_0})^{\otimes N} \\
& (x_1, \dots, x_N) \mapsto ({\pmb \zeta}_1^{(0)}, {\pmb \zeta}_2^{(0)},\dots, {\pmb \zeta}_N^{(0)}). 
\end{split}
\end{gather}
Explicitly, we write $\zeta_{p,i}^{(0)}$ to be the $i$-th component of the ${d_0}$-dimensional vector  ${\pmb \zeta}_p^{(0)}$, for $p\in\{1,\dots,N\}$. 
For instance, $\zeta_{p,i}^{(0)} = f_{i}(x_p)$ for $i\in\{1,2,\dots,2n\}$ in (\ref{def:loc_basis}).

Subsequently, we will have $L$ iterations of feature learning. Each iteration consists of three operations: batch normalisation, convolution, and pooling. 
In what follows we will discuss them individually.

\vspace{7pt}
{\em Batch Normalisation:}
\vspace{2pt}

In q-CNN, we employ a somewhat unusual product pooling to correlate information captured in different spacial locations of the image. This product operation is prone to numerical instability such as overflow and underflow. To remedy this,  we  use batch normalisation to standardise the input features of each layer such that they have a specific (learnable) mean ${\pmb \mu}^{(\ell)}$ and variance ${\pmb \sigma}^{(\ell)}$ over the spacial and batch dimensions. 
In other words, in the $\ell$-th iteration and  for a given batch, let ${\pmb \mu}_b^{(\ell)}$ and ${\pmb \sigma}_b^{(\ell)}$ be the mean and variance of the input ${\pmb \zeta}^{(\ell)}$ over the spatial (denoted by $p$ above) dimensions and over the data points in the batch $b$. Then the batch normalisation layer amounts to the following affine transformation: 
\begin{gather}\label{BN1}\begin{split}
{\rm BN}: ~~ &   ({\mathbb R}^{d_\ell})^{\otimes 2^{L-\ell}}   \to   ({\mathbb R}^{d_\ell})^{\otimes 2^{L-\ell}}  \\ 
& ({\pmb \zeta}_1^{(\ell)}, {\pmb \zeta}_2^{(\ell)},\dots, {\pmb \zeta}_{2^{L-\ell}}^{(\ell)}) \mapsto ({\pmb \zeta'}_1^{(\ell)}, {\pmb \zeta'}_2^{(\ell)},\dots, {\pmb \zeta'}_{2^{L-\ell}}^{(\ell)}),
\end{split}
\end{gather}
with ${\pmb \zeta'}_p^{(\ell)} = {\bf w}^{(\ell)}\, {\pmb \zeta}_p^{(\ell)} +  {\bf z}^{(\ell)}$, or 
\begin{equation}\label{BN}
{\zeta'}_{p,i}^{(\ell)} =w_i^{(\ell)}  {\zeta}_{p,i}^{(\ell)} +z_i^{(\ell)} 
\end{equation}
 in components. In the above, we have that $ {\bf w}^{(\ell)}={\rm diag}(w_1^{(\ell)},\dots,w_{d_\ell}^{(\ell)} )$ is a diagonal matrix and ${\bf z}^{(\ell)}$ is a vector, 
 with entries 
 \begin{equation}
w_i^{(\ell)}:=\frac{\sigma_i^{(\ell)}}{\sqrt{\sigma_{b,i}^2+\epsilon}}~,~~z_i^{(\ell)}:= \mu_i^{(l)} - \frac{\mu_{b,i} \, \sigma_i^{(\ell)}}{\sqrt{\sigma_b^2+\epsilon}}~
\end{equation}
where $\epsilon$ is a small regularizing constant.

\vspace{9pt}
{\em Convolution:}
\vspace{2pt}

We consider a convolution of window size $1\times 1$. Note that this is nevertheless non-trivial since we have $d_\ell>1$ channels. As a result the convolution mixes information carried in different channels though not in different spacial locations. 
Writing the weight tensor in the $\ell$-th layer as a matrix  ${\bf a}^{(\ell)}$ of size $d_{\ell+1} \times d_{\ell}$, we have
\begin{gather}\begin{split}
{\rm Conv}: ~~ &   ({\mathbb R}^{d_\ell})^{\otimes 2^{L-\ell}}   \to  ({\mathbb R}^{d_{\ell+1}})^{\otimes 2^{L-\ell}}  \\ 
& ({\pmb \zeta'}_1^{(\ell)}, {\pmb \zeta'}_2^{(\ell)},\dots, {\pmb \zeta'}_{2^{L-\ell}}^{(\ell)}) \mapsto ({\pmb \xi}_1^{(\ell)}, {\pmb \xi}_2^{(\ell)},\dots, {\pmb \xi}_{2^{L-\ell}}^{(\ell)}),
\end{split}
\end{gather}
with ${\pmb \xi}_p^{(\ell)} = {\bf a}^{(\ell)}\, {\pmb \zeta'}_p^{(\ell)}$, or
\begin{equation}\label{conv_kernel}
{\xi}_{p,j}^{(\ell)} =\sum_i {a}_{ji}^{(\ell)} {\zeta'}_{p,i}^{(\ell)}  
\end{equation}
 in components. 
Note that the tensor ${\bf a}^{(\ell)}$ does not depend on the spacial location $p$, a feature often referred to as weight sharing in the context of the usual CNN architecture.

\vspace{9pt}
{\em Product Pooling:}
\vspace{2pt}

Following each such convolution there is a product pooling operation which perform the same-channel product of the corresponding (one-dimensional) features in non-overlapping spacial windows of size $2$, thus reducing the spacial size of the feature map by a factor of $2$. In other words, we have 
\begin{gather}\begin{split}
{\rm Pool}: ~~ &    ({\mathbb R}^{d_{\ell+1}})^{\otimes 2^{L-\ell}}  \to  ({\mathbb R}^{d_{\ell+1}})^{\otimes {2^{L-\ell-1}}}  \\ 
&  ({\pmb \xi}_1^{(\ell)}, {\pmb \xi}_2^{(\ell)},\dots, {\pmb \xi}_{2^{L-\ell}}^{(\ell)})\mapsto ({\pmb \zeta}_1^{(\ell+1)}, {\pmb \zeta}_2^{(\ell+1)},\dots, {\pmb \zeta}_{2^{L-\ell-1}}^{(\ell+1)}),
\end{split}
\end{gather}
where 
\begin{equation}\label{delta_pool}
 \zeta^{(\ell+1)}_{p,i}= \xi^{(\ell)}_{2p-1,i}    \xi^{(\ell)}_{2p,i} = \sum_{j,k} \delta_{i,j,k} \xi^{(\ell)}_{2p-1,j}    \xi^{(\ell)}_{2p,k}  .
\end{equation}

\vspace{9pt}
{\em Classification:}
\vspace{2pt}

The last pooling layer is followed by a batch normalisation operation and a dense linear layer (trivially a convolution on a $1\times1$ feature map), which maps the remaining features to the classification space of $|\mathcal{C}|$ labels,
\begin{gather}\begin{split}
{\rm Cl:} ~~ &   {\mathbb R}^{d_L}   \to  {\mathbb R}^{|\mathcal{C}|}  \\ 
& {\pmb \zeta'}_1^{(L)} \mapsto {\pmb \alpha},
\end{split}
\end{gather}
with ${\pmb \alpha} = {\bf a}^{(L)}\, {\pmb \zeta'}_1^{(L)}$, or
\begin{equation}
{\alpha}_{y} =\sum_{i} {a}_{yi}^{(L)} {\zeta'}_{1,i}^{(L)}  
\end{equation}
in components.

\begin{figure}[ht]
\label{cong_fig}
\centering
\includegraphics[scale=0.5]{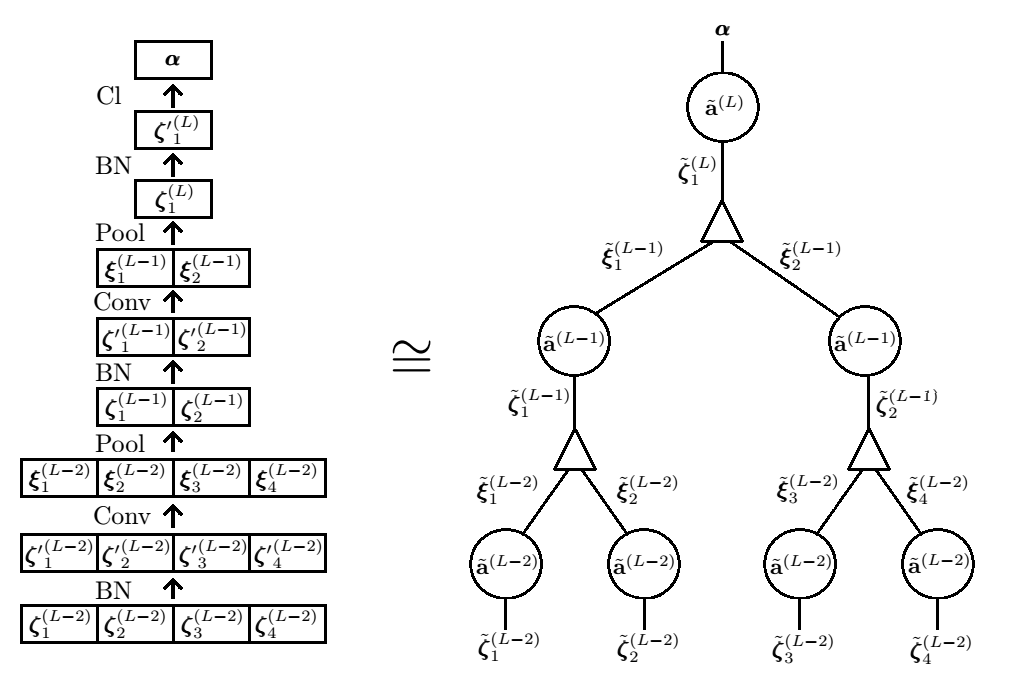}
\caption{The neural network architecture used in this paper (left), and its tensorial description (right). The triangles represent the delta tensor implementing the same-channel product pooling (\ref{delta_pool}).}
\end{figure}

\subsection{Tensor Description} 
As mentioned before, without batch normalisation, the architecture consisting of the specific form of convolutional and product pooling layers described above is the convolutional arithmetic circuit that the authors proposed in \cite{cohen2016expressive}, and further studied in \cite{levine2017deep,levine2019quantum}.  It was also pointed out that such an architecture implements a (hierarchical) Tucker decomposition of the network tensors.  
To have a trainable network we additionally apply batch normalisation. Naively, this destroys the description of the network as a tensor operation, since batch normalisation (\ref{BN}) is an affine instead of a linear transformation.  However, this can be easily remedied by adding an additional dimension, corresponding to the ``constant term'', in all layers. Concretely, we can equivalently describe the affine transformation (\ref{BN1}) as the following linear transformation: 
\begin{gather}\label{BN3}\begin{split}
{\rm BN}: ~~ &   ({\mathbb R}^{d_\ell+1})^{\otimes 2^{L-\ell}}   \to   ({\mathbb R}^{d_\ell+1})^{\otimes 2^{L-\ell}}  \\ 
& (\tilde{\pmb \zeta}_1^{(\ell)}, \tilde{\pmb \zeta}_2^{(\ell)},\dots, \tilde{\pmb \zeta}_{2^{L-\ell}}^{(\ell)}) \mapsto (\tilde{\pmb \zeta'}_1^{(\ell)}, \tilde{\pmb \zeta'}_2^{(\ell)},\dots, \tilde{\pmb \zeta'}_{2^{L-\ell}}^{(\ell)}),
\end{split}
\end{gather}
where 
\begin{equation}
  \tilde{\pmb \zeta}_p^{(\ell)}  = \begin{pmatrix} 1 \\ {\pmb \zeta}_p^{(\ell)}
  \end{pmatrix}, ~~  \tilde{\pmb \zeta'}_p^{(\ell)}  = \begin{pmatrix} 1 \\ {\pmb \zeta'}_p^{(\ell)}
  \end{pmatrix},
\end{equation}
and 
$\tilde{\pmb \zeta'}_p^{(\ell)} = \tilde{\bf w}^{(\ell)}\, \tilde{\pmb \zeta}_p^{(\ell)}$, with
\begin{equation}
    \tilde{\bf w}^{(\ell)} = \begin{pmatrix}
    1 & 0_{1\times d_\ell} \\ {\bf z} & {\bf w}^{(\ell)}.
     \end{pmatrix}
\end{equation}

At the same time, the convolution and product pooling layers can also be described in terms of the $(d+1)$-dimensional vectors $\tilde{\pmb \zeta}_p^{(\ell)}$, $\tilde{\pmb \zeta'}_p^{(\ell)}$ and
$\tilde{\pmb \xi}_p^{(\ell)}$ in a straightforward way.
For instance, the batch normalisation and the convolutional layers can be described as a combined tensorial operation: 
\begin{gather}\begin{split}
{\rm BN + Conv}: ~~ &   ({\mathbb R}^{d_\ell+1})^{\otimes 2^{L-\ell}}   \to  ({\mathbb R}^{d_{\ell+1}+1})^{\otimes 2^{L-\ell}}  \\ 
& (\tilde {\pmb \zeta}_1^{(\ell)}, \tilde{\pmb \zeta}_2^{(\ell)},\dots, \tilde{\pmb \zeta}_{2^{L-\ell}}^{(\ell)}) \mapsto (\tilde{\pmb \xi}_1^{(\ell)}, \tilde{\pmb \xi}_2^{(\ell)},\dots, {\pmb \xi}_{2^{L-\ell}}^{(\ell)}). 
\end{split}
\end{gather}
with $\tilde{\pmb \xi}_p^{(\ell)} = \tilde{\bf a}^{(\ell)}\, \tilde{\pmb \zeta'}_p^{(\ell)}$, where
\begin{equation}\label{conv_kernel_bn}
\tilde{\bf a}^{(\ell)}= \begin{pmatrix}
    1 & 0_{1\times d_\ell} \\{\bf a}^{(\ell)} {\bf z}& {\bf a}^{(\ell)}{\bf w}^{(\ell)}.\end{pmatrix}
\end{equation}
for $\ell=0,1,\dots,L-1$. 
In the final classification layer, we simply have 
\begin{equation}\label{conv_kernel_CL}
\tilde{\bf a}^{(L)}= \begin{pmatrix}
{\bf a}^{(\ell)} {\bf z}& {\bf a}^{(\ell)}{\bf w}^{(\ell)}\end{pmatrix} .
\end{equation}
since we do not need the additional constant channel in the final output. 

As a result, the q-CNN we defined above can also be described in terms of tensor networks, as we will further describe in the following section.

\section{Quantum Properties}
In this section we discuss the description the data and the q-CNN network in the language of quantum many-body systems, which then enables the definition and calculation of the entanglement entropy of the network states. 

\subsection{A Quantum Description}
\label{quantum_description}

To describe the neural network introduced in \S\ref{sec:architecture}, let us first describe our quantum Hilbert space in terms of a space of $L^2$-integrable complex functions\footnote{In this current work we will only use real functions in practice, though the generalisation  allowing for  complex functions is straightforward, at least in theory.}. 
Explicitly, consider 
\begin{equation}
L^2(S^1,{\mathbb C}):= \{f: [0,1]\to \CC ~\lvert~ f(0)=f(1), \int_0^1 dx\,  \overline{f(x)} f(x)  ~~\text{exists} \}. 
\end{equation}
As is well-known, this space is equipped with a natural norm 
\begin{equation}\label{dfn:norm}
    \langle f,g \rangle : = \int_0^1 dx\,  \overline{f(x)} g(x) , 
\end{equation}
and an orthonormal basis is given by the Fourier basis\footnote{Apart from the Fourier basis, other bases may be used as well.  Concretely, we have implemented network with the Legendre basis for $L^2\left([-1,1]\right)$, with which we obtain similar results in terms of accuracies.}
\begin{equation}\label{def:loc_basis}
    f_0(x) = 1, ~~ f_{2k-1}(x) = \sqrt{2} \cos(2\pi k x) ,   ~~ f_{2k}(x) = \sqrt{2} \sin(2\pi k x) 
\end{equation}
for $k\in \mathbb N$.
To obtain a finite-dimensional Hilbert space, we restrict to a subspace by truncating the modes with frequency larger than a given $n\in \mathbb N$: 
\begin{equation}
L^2_n(S^1):= \{f\in L^2(S^1,{\mathbb C}) \lvert \langle f, f_{k} \rangle =0 ~~\forall~ k> 2n\}. 
\end{equation}

We will identify this space with the our {\em local Hilbert space} 
\begin{equation}
{\cal H}_{\rm loc}\cong L^2_n(S^1) \cong {\mathbb C}^{2n+1}
\end{equation}
describing the state of the local pixel (or spin in the physics analogy). Explicitly, corresponding to the orthonormal basis (\ref{def:loc_basis}) for $L^2_n(S^1)$, we introduce an orthonormal basis $\vert f_0\rangle, \dots, \vert f_{2n}\rangle$ for ${\cal H}_{\rm loc}$, and it follows from the orthonormality that 
\[\sum_{i=0}^{2n} |f_i\rangle\langle f_i| = {\bf 1}_{{\cal H}_{\rm loc}}. 
\]
 We also introduce $\langle x \lvert \in {\cal H}_{\rm loc}^\ast$ to be giving the {\em evaluation map}:
\begin{equation}
    \langle x \lvert f \rangle = {\bf ev}(f,x) = f(x). 
\end{equation}
In other words, we can think of $|x\rangle$ as the position eigenstates, and $f(x)$  can be thought of as the corresponding wavefunction associated with the state $\lvert f \rangle$. 
Explicitly, one has 
\begin{equation}
\langle x \lvert\, =\sum_{i=0}^{2k}  f_i(x) \langle f_i \lvert~, 
\end{equation}
and 
\begin{equation}
\int_0^1 dx\, \lvert x \rangle \langle x \lvert = \sum_{i=0}^{2n} |f_i\rangle\langle f_i| = {\bf 1}_{{\cal H}_{\rm loc}}~.
\end{equation}

Note also that 
\begin{equation}
\langle x\lvert x'\rangle=D_n(x-x')~,~~D_n(z)=1+2\sum_{k=1}^n \cos(2\pi kz)~,  
\end{equation}
is the so-called Dirichlet kernel on $L_n^2(S^1)$, and has the Dirac delta function $\delta(x)$ as the limit when $n\to \infty$.

Now, consider a system with $N$ pixels (or lattice sites in the physics analogy), and we have 
\begin{equation}\label{dfn:Hilbertspace}
   {\cal H} =  {\cal H}_{\rm loc}^{\otimes N} \cong L^2_n(T^N)\cong (L^2_n(S^1))^{\otimes N}. 
\end{equation}

We will write the coordinates ${\bf x} = (x_1,\dots, x_N)$ for the $N$-torus $T^N = [0,1)^{\otimes N}$, and introduce the following natural (orthonormal) basis for $ {\cal H}$, 
\begin{equation}\label{def:orthonormalBasis}
\lvert f_{i_1\dots i_N} \rangle =\bigotimes_{p=1}^N \lvert f_{i_p}\rangle_p, ~~~~ i_p \in \{0,1,\dots,2n\},
\end{equation}
corresponding to the basis $\lvert f_{i}\rangle_p$ for the local Hilbert space of $p$-th pixel.

Identifying a greyscale image with $N$ pixels with a point in $[0,1]^{\otimes N}$, 
the ``position eigenstate'' $\lvert {\bf x} \rangle$ corresponds in our case to a specific image. 
Note that, by working with the space (\ref{dfn:Hilbertspace}), we restrict ourselves to a periodic representation that is invariant under the action of swapping a zero with an one \footnote{but not invariant under swapping 0.05 and 0.95, For this reason the periodicity is not so significant in practice for real data. The choice for a periodic representation is simply made for convenience and does not seem to hinder the classification in our experiments. }.
Just as in the single pixel/particle case, its expansion in the orthonormal basis (\ref{def:orthonormalBasis}) reads
\begin{equation}
\lvert {\bf x} \rangle = \sum_{{\bf i} \in \{0,1,\dots, 2n\}^{\otimes N}} \Psi_{\bf i}({\bf x}) \lvert f_{\bf i} \rangle
\end{equation}
where we have written ${\bf i} = (i_1,\dots,i_N)$, and   the coefficient function reads\footnote{Note that $\lvert {\bf x} \rangle$ is a product state and is in particular not entangled.}
\begin{equation}
\Psi_{\bf i}({\bf x}) = \prod_{p=1}^N f_{i_p}(x_p).
\end{equation}

It then follows immediately that the final score function output from our neural network can be viewed as the value of the wavefunction corresponding to  the ``network state'' \footnote{Note that a wavefunction is complex-valued in general. As mentioned earlier, here we work with real functions   for convenience and have $\Psi_{y;{\bf i}}^\ast=\Psi_{y;{\bf i}}$.  }
\begin{equation}
\lvert \Psi_y  \rangle  =\sum_{{\bf i}\in \{0,1,\dots, 2n\}^{\otimes N}} \Psi_{y;{\bf i}}\lvert f_{\bf i} \rangle
\end{equation}
 corresponding to the label $y\in \mathcal{C}$: 
\begin{equation}\label{wavefn}
\alpha_y({\bf x}) := \sum_{{\bf i} \in \{0,1,\dots, 2n\}^{\otimes N}} \Psi_{\bf i}({\bf x})\, \Psi_{y;{\bf i}}    
= \langle {\bf x} \lvert \Psi_y\rangle.
\end{equation}

At this point, it is tempting to associate to our wavefunction the usual probabilitic interpretation \`a la Born's rule. 
To do this, introduce the normalised network states 
\begin{equation}\label{normalisation_wavefn}
\lvert \Psi_{y,0}  \rangle := {\frac{1}{\sqrt{\sum_{y\in {\cal C}}   \langle\Psi_y  \lvert \Psi_y  \rangle }}}\,\lvert \Psi_y  \rangle
\end{equation}
and define the joint probability density function 
\begin{equation}
p(y,\textbf{x}):= \lvert \langle {\bf x}\lvert \Psi_{y,0}  \rangle \lvert^2 . 
\end{equation}

Building a (generative) network learning the states $\lvert \Psi_{y,0}  \rangle$ that gives a good approximation to the above probability density function is beyond the scope of the current paper; here we focus on the classification task and hence we can only trust $p(y,\textbf{x})$ in the subspace of ${\cal H}$ where $\textbf{x}$ resembles the training data in some way. 
Instead, for the classification task at hand, the relevant probability is the conditional  probability 
\begin{equation}
\label{cond_prob}
p(y|\textbf{x})=\frac{p(y,\textbf{x})}{p(\textbf{x})}=\frac{\lvert \langle {\bf x}\lvert \Psi_{y,0}  \rangle \lvert^2 }{\sum_{y'\in {\cal C}}\lvert \langle {\bf x}\lvert \Psi_{y',0}  \rangle \lvert^2 } =\frac{\lvert \langle {\bf x}\lvert \Psi_{y}  \rangle \lvert^2 }{\sum_{y'\in {\cal C}}\lvert \langle {\bf x}\lvert \Psi_{y'}  \rangle \lvert^2 } ~,
\end{equation}
As is manifest from the last equality, the conditional probability is insensitive to the normalisation (\ref{normalisation_wavefn}) of the network state $\lvert \Psi_y\rangle$. This justifies our classification given the network output $\alpha_y({\bf x}) =\langle {\bf x} \lvert \Psi_y\rangle$:  
\begin{equation}\label{label1}
y({\bf x}) := \argmax_{y} \lvert\alpha_y({\bf x})  \lvert. 
\end{equation}
Note that this is different from the more common ways of assigning probabilities to the outputs of such a classification network, such as through a softmax function.

\subsection{Entanglement entropy}
\label{EE_theory}

Equipped with the interpretation described in \S\ref{quantum_description} of our architecture as quantum states, we are now ready to ``measure'' the behaviour of the the neural network using quantum mechanical tools. 
In particular, in this subsection we will discuss the entanglement entropy of the network. 

As the name suggests, the entanglement entropy measures the extent to which a quantum state is entangled across two different subsystems,  $\mathtt{U}$ and $\overline{\mathtt{U}}$. It is defined as the von Neumann entropy of the reduced density matrix. For instance, if the state is a product state of two separate systems,  then its entanglement entropy (with respect to the subsystems) vanishes.

Here, we consider a bipartition $(\mathtt{U},\overline{\mathtt{U}})$ of the images that spatially splits the pixels  $\textbf{x}=\left(\textbf{x}_\mathtt{U}, \textbf{x}_{\overline{\mathtt{U}}}\right)$ into two groups, with the corresponding Hilbert space decomposition ${\cal H} = {\cal H}_\mathtt{U}\otimes   {\cal H}_{\overline{\mathtt{U}}}$. 
To discuss entanglement entropy of different network states corresponding to different classification labels $y\in {\cal C}$, we normalise the network state as
\begin{equation}
\lvert \Psi_{y,\ast}  \rangle := {\frac{1}{\sqrt{\langle\Psi_y  \lvert \Psi_y  \rangle}} }\,\lvert \Psi_y  \rangle.
\end{equation}
Note that this is a different normalisation from (\ref{normalisation_wavefn}). 
The corresponding reduced density matrix, obtained by tracing the density matrix over the subspace ${\cal H}_{\overline{\mathtt{U}}}$,  reads 
\begin{equation}\label{def:RDM}
\rho_{y;\mathtt{U}} := \Tr_{{\cal H}_{\overline{\mathtt{U}}}}  \left[| \Psi_{y,\ast} \rangle\langle \Psi_{y,\ast} |\right]~, 
\end{equation}
which satisfies $\Tr_{{\cal H}_\mathtt{U}} \rho_{y;\mathtt{U}}  =\langle\Psi_{y,\ast} |\Psi_{y,\ast}\rangle = 1 $.
Then the entanglement entropy of the network state 
$\lvert \Psi_{y,\ast}  \rangle$ corresponding to the bipartition $(\mathtt{U},\overline{\mathtt{U}})$ is given by 
\begin{equation}\label{dfn:ee}
	{\cal S}(\rho_{y;\mathtt{U}})  = -\Tr_{{\cal H}_\mathtt{U}} \left(\rho_{y;\mathtt{U}} \log \rho_{y;\mathtt{U}}\right).
\end{equation}
This quantity measures the extent to which the quantum state is entangled across the subspaces ${\mathtt{U}}$ and $\overline{\mathtt{U}}$, i.e. the degree to which the quantum state fails to be separable into two parts, belonging to two subsystems. 
 For instance, if $|\Psi_y\rangle$ is a product state $|\Psi_y\rangle = |\Psi_y\rangle_{\mathtt{U}} \otimes |\Psi_y\rangle_{\overline{\mathtt{U}}}$ with $|\Psi_y\rangle_{\mathtt{U}} \in 
{\cal H}_{\mathtt{U}}$ and $|\Psi_y\rangle_{\bar{\mathtt{U}}} \in {\cal H}_{\bar{\mathtt{U}}}$, then the probability (\ref{cond_prob}) also takes the form of a product of contribution from $\mathtt{U}$ and $\bar{\mathtt{U}}$, and we have ${\cal S}(\rho_{y;\mathtt{U}})=0$.

In what follows we will focus on the bipartition with $\mathtt{U}$ and $\overline{\mathtt{U}}$ each contains $N/2$ pixels (or lattice sites), corresponding to the  two inputs of the top pooling tensor. 
In the next section we flatten the input images in ways such that this bipartition corresponds to either the left/right or the up/down separation of the images. 
In this case, analogous to the basis (\ref{def:orthonormalBasis}) for the total Hilbert space ${\cal H}$, we introduce the corresponding orthonormal basis for ${\cal H}_{\mathtt U}\cong{\cal H}_{\overline{\mathtt U}}\cong (\CC^{2n+1})^{\otimes N/2} $ : 
\begin{equation}
\lvert f_{\bf i} \rangle = \lvert f_{i_1}\rangle \otimes \dots \otimes \lvert f_{i_{\frac{N}{2}}}\rangle, 
\end{equation}
for ${\bf i} = (i_1,\dots, i_{\frac{N}{2}}) \in \{0,1,\dots,2n\}^{\otimes {\frac{N}{2}}}$.  
The top pooling layer of the tensor decomposition gives   the following expression 
\begin{equation}
|\Psi_y\rangle=\sum_{I} \tilde {\bf a}^{(L)}_{yI} ~|\phi_I\rangle_\mathtt{U}\otimes |\phi_I\rangle_{\overline{\mathtt{U}}}~ , 
\end{equation}
where
\begin{equation}
\lvert \phi_I\rangle = \sum_{{\bf i}  \in \{0,1,\dots,2n\}^{\otimes {\frac{N}{2}}}} \phi_{I,{\bf i}} \lvert f_{\bf i} \rangle
\end{equation}
for both $|\phi_I\rangle_\mathtt{U}$ and $|\phi_I\rangle_{\overline{\mathtt{U}}}$. Recall that $|\phi_I\rangle_\mathtt{U}= |\phi_I\rangle_{\overline{\mathtt{U}}}$ under the natural isomorphism ${\cal H}_{\mathtt U} \cong {\cal H}_{\bar{\mathtt U}}$, and this is a consequence of  the weight sharing feature of  our architecture. Also note that the $y$-dependence in $|\Psi_y\rangle$ comes entirely from the top layer tensor $ {\bf a}^{(L)}$.

Putting it together, we have the following expression for the reduced density matrix (\ref{def:RDM})
\begin{gather}
\begin{split}
\rho_{y;{\mathtt{U}}} &= {\frac{1}{\langle\Psi_y  \lvert \Psi_y  \rangle}} \sum_{I, J}  (\tilde a^{(L)}_{y J})^\ast\, \tilde a^{(L)}_{y I}
\Tr_{{\cal H}_{\overline{\mathtt{U}}}}\left(|\phi_I\rangle_{\overline{\mathtt{U}}}\otimes |\phi_I\rangle_{{\mathtt{U}}} \;{ }_{{\mathtt{U}}} \langle \phi_J \lvert \otimes  \;{ }_{\overline{\mathtt{U}}} \langle \phi_J \lvert \right)\\
&=  {\frac{1}{\langle\Psi_y  \lvert \Psi_y  \rangle}}  \sum_{{\bf i}, {\bf i}'   \in \{0,1,\dots,2n\}^{\otimes {\frac{N}{2}}}} (M_y)_{{\bf i}, {\bf i}'} \lvert f_{\bf i} \rangle  \langle  f_{{\bf i}'} \lvert 
\end{split}
\end{gather}
where the matrix elements are given by 
\begin{equation}
	(M_y)_{{\bf i}, {\bf i}'} = \sum_{I,J}\sum_{{\bf i}'' \in \{0,1,\dots,2n\}^{\otimes {\frac{N}{2}}}} 
	  (\tilde a^{(L)}_{yJ})^\ast \,\tilde a^{(L)}_{yI} 
	  \phi_{I,{\bf i}''} \, \phi_{J,{\bf i}''}^\ast \,
	   \phi_{I,{\bf i}} \, \phi_{J,{\bf i}'}^\ast .  
\end{equation}

From the eigenvalues $m_{\bf i}$, ${\bf i} \in \{0,1,\dots,2n\}^{\otimes {\frac{N}{2}}}$ of the above matrix,  we readily compute the entanglement entropy (\ref{dfn:ee}) of the network states to be
\begin{equation}
		{\cal S}(\rho_{y;\mathtt{U}}) = -\sum_{{\bf i} \in \{0,1,\dots,2n\}^{\otimes {\frac{N}{2}}}} p_{\bf i} \log p_{\bf i}, 
\end{equation}
with 
\begin{equation}
p_{\bf i} := \frac{m_{\bf i}}{\sum_{\bf i'} m_{\bf i'} }.
\end{equation}

\section{Experiments}

\subsection{Setup}
We trained the network on two datasets, MNIST \cite{lecun-mnisthandwrittendigit-2010} and Fashion-MNIST \cite{xiao2017fashion}, using PyTorch \cite{paszke2019pytorch}. Each contains 60000 training images, 10000 test images, and a total number of $|\mathcal{C}|=10$  classes. The images are resized to $16\times16$, so that the flattened array contains $N=256=2^8$ elements, resulting in a depth $L=8$ network.
 
The number of parameters of the network can easily be calculated,
\begin{equation}
P\left(N;\lbrace d_\ell\rbrace\right)=\sum_{l=0}^{L=\log_2 N}\left( d_\ell  d_{\ell+1}+2 d_\ell\right)~,
\end{equation}
where the first term comes from the convolution (without bias), and the second from the batch normalisation layers. Note that, in the case that $d_\ell$ is independent of $\ell$, the  number of parameters grows like $\log_2 N$ with $N$. 

Our main examples will be 
 networks  with number of channels increasing with the depth, 
 $d_\ell=d(\ell+1)$ for $\ell=0,\dots,L$, and  $d_{L+1}=|\mathcal{C}|=10$ for the final layer. 
 This results in the following total number of parameters
\begin{equation}
\begin{split}
&P(N)=\frac{d^2}{3}\left(\log_2 N\right)^3 + (d^2+d)\left(\log_2 N\right)^2 + \left(\frac{2}{3}d^2+13d\right)\log_2 N+12d~, \\
&P(2^8)=240d^2+180d~,
\end{split}
\end{equation}
which scales quadratically with $d$ and merely with (powers of) $\log_2N$, a result of  the tree tensor network structure, which eliminates quadratic growth,  as well as weight-sharing, which eliminates linear growth with $N$ . We used values of $d$ from $2$ up to $40$, resulting in a number of parameters in the range $(1320, 391200)$.

Classification of an input $\textbf{x}$ is done by choosing the label $y$ for which the output $|\alpha_y(\textbf{x})|$ is the largest (cf. (\ref{label1})), consistent with the quantum  interpretation discussed in section \ref{quantum_description}. 
We used the following  square distance loss function, 
\begin{equation}
\label{cost}
\mathcal{L}=\frac{1}{|\mathcal{B}|}\sum_{b=1}^{|\mathcal{B}|}\sum_{y=1}^{|\mathcal{C}|}\left( \alpha_y(\textbf{x})-\delta_{y,\mathfrak{l}(\textbf{x})} \right)^2~,
\end{equation}
where $\mathfrak{l}(\textbf{x})$ is the correct label of $\textbf{x}$. Reducing this cost means that the output in the correct label channel moves closer to $1$, while the outputs in the rest of the channels move closer to $0$ \footnote{
Given (\ref{cond_prob}) and (\ref{label1}) it might be tempting to use the loss function $|\alpha_y(\textbf{x})|$ or $|\alpha_y(\textbf{x})|^2$, or even just $-\log\left(\frac{|\alpha_{\mathfrak{l}(\textbf{x})}(\textbf{x})|^2}{\sum_y |\alpha_y(\textbf{x})|^2}\right)$. However, these choices all give worse training results than (\ref{cost}). One simple reason is just that the above loss is invariant under a sign change of $\alpha_y$ and hence does not admit a unique minimum.}.

Optimization was done by using the AdamW optimizer \cite{loshchilov2017decoupled}, with weight decay parameter $0.01$ and learning rate parameter $0.01$, which was reduced by half every 10 epochs. Batch size was chosen equal to $50$ and we trained for a total of 90 epochs.
Tensor network computations for evaluating the entanglement entropy were done using the Tensornetwork library \cite{roberts2019tensornetwork}. The corresponding code can be found \href{https://github.com/vanagia/entangled-q-cnn}{here}.

\subsection{Results and discussion}

\begin{figure}[ht]
\label{fig1}
\centering
\includegraphics[scale=0.69]{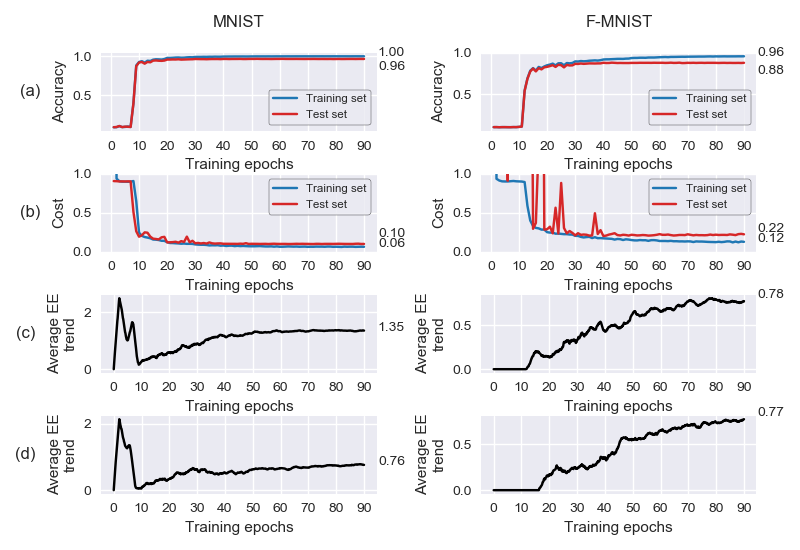}
\caption{(a), (b) Accuracy and cost during training for a typical  run of an experiment with $d=18$ for both datasets. (c) The corresponding left/right average entanglement entropy trend, where we average over the classes with an averaging window of 2 epochs. (d) The up/down average entanglement entropy trend for a different run with the same hyperparameters.\label{Fig:experiment}}
\end{figure}

The best  (average) test accuracies achieved among all experiments were $97\%$ for MNIST and $89\%$ for Fahsion-MNIST, which are comparable to the results obtained in \cite{efthymiou2019tensornetwork}.  Note that we did not perform any exhaustive hyperparameter search, so it is likely that the achieved accuracies can be improved further.
Figure \ref{Fig:experiment} depicts a typical (as opposed to optimal in terms of attained accuracies) run of an experiment with $d=18$ ($81000$ parameters), for both datasets. Notice that after epochs $30$-$50$ the test accuracy and cost value start to slowly approach their final values.
Figure  \ref{Fig:experiment}(c) shows the trend (averaged over two epochs) of the average entanglement entropy  of all $10$ output channels, according to a left/right partition, as it develops during training.
Upon initialization of the network the EE is practically zero, and during the first few epochs its value can vary widely among different initialization seeds, optimizers and number of hyperparameters. However, after the accuracy and the cost value has stopped changing rapidly, which happens after around 10 epochs, the EE starts to increase steadily. This signifies a rise in the degree of correlations that are built between the left/right parts of the network in order to  classify the images with more precision. This is to be contrasted with  the ``accidental'' entanglement that appear during the first $10$ epochs in the MNIST case, when the network is not yet able to correctly classify the images systematically.

\begin{figure}[ht]
\label{fig2}
\centering
\includegraphics[scale=0.69]{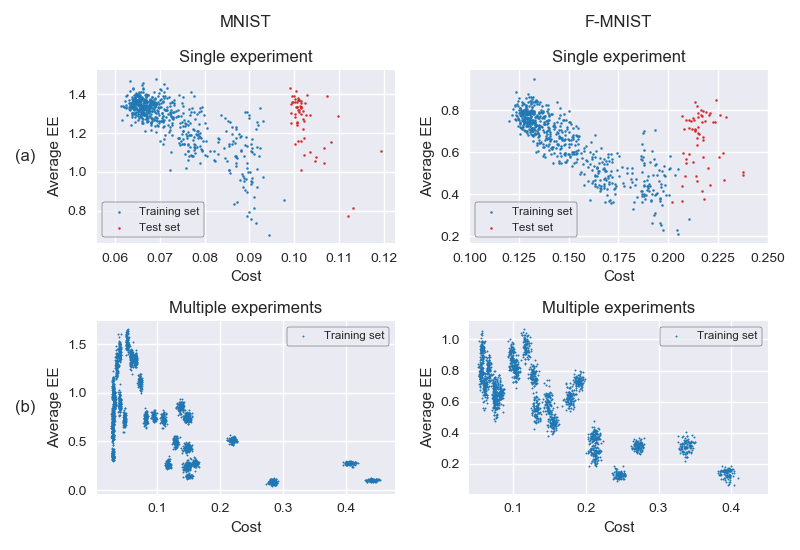}
\caption{(a) The average entanglement entropy trend versus the value of the cost function for a typical run of an experiment with $d=18$, for both datasets. (b) The corresponding plots across multiple experiments.\label{Fig:EE}}
\end{figure}

Note that in the MNIST case the value of the EE seems to stabilise during the final epochs, whereas it appears to keep growing in the case of F-MNIST. A possible explanation is that at an accuracy of $\sim 88\%$ there are many more misclassifications in the latter case compared to the former case, and thus the minimum EE needed for near-optimal classification has not yet been reached. In support of this, we note that after around epoch $30$ the F-MNIST plot resembles the region between epochs $10$ to $40$ of the MNIST plot, where the corresponding accuracy is still lower than the final achieved value.

Figure  \ref{Fig:experiment}(d) shows the average EE trend for a different run with the same hyperparameters, but with a different flattening of the images such that the top pooling corresponds to an up/down division of the images. 
Note that the development trend of the up/down EE is very similar to that of the left/right EE. We believe that the heuristic explanation of the trend mentioned above for the left/right EE also applies to the case of up/down EE.

As mentioned before, in \cite{liu2019machine} the (final value of the)  entanglement entropy of a rather different neural network has been measured in the context of MNIST dataset. We note that the order of magnitude of the final EE we measured is similar to that in \cite{liu2019machine}, suggesting that EE is indeed a robust quantitative measure of certain key properties of the given tasks.

Finally, in Figure \ref{Fig:EE}(a) we depict the average EE versus the value of the cost function, for the same run throughout the training process, as recorded in Figure \ref{Fig:experiment}(a,b,c).
The data before the thirtieth epoch are discarded due to their large fluctuations, as noted earlier.
We observe a distinct negative correlation between EE and the value of the cost function, which is to say that higher values of the EE tend to appear when the network has lower values for the cost function. 
This can be seen more easily in Figure \ref{Fig:EE}(b), where the plots contain data from many different experiments, across different initialization seeds and choices of hyperparameters, 
which appear as ``islands'' in a larger landscape that also exhibits some degree of negative correlation.
This is consistent with our interpretation that the EE of the network starts increasing steadily only as the network starts learning the finer features of the data.
Also note that this negative correlation is less evident in the test cost; we believe that this is due to the fact that the test cost drops much slower than the training cost after a certain point during training. This suggests that the aforementioned increase of the EE is in part also tied to overfitting.

It would also be interesting to investigate the entanglement structure of a generative network approximating the probability density function (\ref{cond_prob}), as such a network will have a more global knowledge of the Hilbert space $\cal H$.

\section*{Acknowledgements}
We thank Gabriele Cesa, Ying-Jer Kao, Max Welling, and in particular Tolya Dymarsky and Vasily Pestun for helpful conversations. The work of V.A. is supported by ERC starting grant H2020 \# 640159. 
The work of M.C. is supported by ERC starting grant H2020 \# 640159 and NWO vidi grant (number 016.Vidi.189.182). 

\bibliographystyle{unsrt}
\bibliography{refs}

\end{document}